\def\year{2020}\relax
\pdfoutput=1
\documentclass[letterpaper]{article} 
\usepackage{aaai20}  
\usepackage{times}  
\usepackage{helvet} 
\usepackage{courier}  
\usepackage[hyphens]{url}  
\usepackage{graphicx} 
\usepackage{graphicx}  
\usepackage{amsthm}
\theoremstyle{definition}
\newtheorem{example}{Example}[section]
\usepackage[export]{adjustbox}
\usepackage{dsfont}

\nocopyright
\title{RuDaS: Synthetic Datasets for Rule Learning and Evaluation Tools}

\author{
Cristina Cornelio \\
IBM Research\\
cor@zurich.ibm.com
\And
Veronika Thost\\
IBM Research\\
veronika.thost@ibm.com
}

\usepackage{amsmath,amsfonts,bm}

\def\eqref#1{equation~\ref{#1}}

\def\1{\bm{1}}

\DeclareMathAlphabet{\mathsfit}{\encodingdefault}{\sfdefault}{m}{sl}
\SetMathAlphabet{\mathsfit}{bold}{\encodingdefault}{\sfdefault}{bx}{n}

\usepackage{booktabs}
\usepackage{xspace}
\usepackage{subfigure}
\usepackage{verbatim}
\usepackage{listings}
\usepackage{todonotes}
\usepackage{tikz}
\usepackage{pgfplots}
\usetikzlibrary{arrows}
\usetikzlibrary{arrows.meta}

\newcommand{\support}{\text{support}\xspace}
\newcommand{\tool}{\text{RuDaS}\xspace}
\newcommand{\nowa}{\ensuremath{n_{\text{OW}}}\xspace}
\newcommand{\nnoiseadd}{\ensuremath{n_{\text{Noise+}}}\xspace}
\newcommand{\nnoisemiss}{\ensuremath{n_{\text{Noise-}}}\xspace}
\newcommand{\nnoise}{\ensuremath{n_{\text{Noise*}}}\xspace}

\newcommand{\ndag}{\ensuremath{n_{\text{DG}}}\xspace}
\newcommand{\nnskip}{\ensuremath{n_{\text{Skip}}}\xspace}
\newcommand{\db}{\ensuremath{\mathbb{D}}\xspace}
\newcommand{\dbowa}{\ensuremath{\db_{\text{OW}}}\xspace}
\newcommand{\dbowan}{\ensuremath{\db_{\text{OW+Noise}}}\xspace}
\newcommand{\dbn}{\ensuremath{\db_{\text{Noise}}}\xspace}
\newcommand{\sfacts}{\ensuremath{\mathbb{S}}\xspace}
\newcommand{\cfacts}{\ensuremath{\mathbb{C}}\xspace}
\newcommand{\tfacts}{\ensuremath{\mathbb{T}}\xspace}
\newcommand{\ass}{\text{ :- }}
\newcommand{\expred}[1]{\textrm{#1}\xspace}
\newcommand{\exconst}[1]{\textrm{#1}\xspace}
\usepackage{multirow}
\newcommand{\citet}[1]{\citeauthor{#1} \shortcite{#1}}
\newcommand{\citep}{\cite}

\begin{document}

\maketitle

\begin{abstract}
Logical rules are a popular knowledge representation language in many domains, representing background knowledge and encoding information that can be derived from given facts in a compact form.
However, rule 
formulation is a complex process that requires deep domain expertise,
and is further challenged by today's often 
large, heterogeneous, and incomplete 
knowledge graphs. 
Several approaches for learning rules automatically, given 
a set of input example facts,
have been proposed over time,
including, more recently, neural systems.
Yet, the area is missing adequate datasets and evaluation approaches: existing datasets 
often resemble toy examples that neither cover the various kinds of dependencies between rules nor allow for testing scalability.
We present a tool for generating different kinds of 
datasets and for evaluating rule learning systems, including new performance measures.
\end{abstract}

\section{Introduction}

Logical rules are a popular knowledge representation language in many domains. They represent 
domain knowledge, encode information that can be derived from given facts in a compact form, and allow for logical reasoning.
For example, given facts $\expred{parent}(\exconst{ann},\exconst{bob})$ and $\expred{parent}(\exconst{bob},\exconst{dan})$, the \emph{datalog rule} \cite{datalog} 
$$\expred{grandparent}(X,Z)\ass\expred{parent}(X,Y),\expred{parent}(Y,Z)$$ encodes the fact $\expred{grandparent}(\exconst{ann},\exconst{dan})$ and describes its dependency on the other facts. Moreover, if the data grows and new facts are added, we can automatically derive new knowledge.
Since rule formulation is complex and requires domain expertise,
\emph{rule learning} \cite{Raedt-08-Logical-and-relational-learning,Fuernkranz+-12-Foundations-of-Rule-Learning} 
has been an area of active research in AI for a long time, also under the name \emph{inductive logic programming} (ILP). 
It has recently revived with the 
increasing use of knowledge graphs (KGs), which can be considered as large fact collections. 
KGs are used in various domains such as in the Semantic Web or with companies such as Google \cite{google} or Amazon \cite{amazon}, and there are large knowledge bases in the medical domain. Useful rules over these knowledge bases would obviously provide various benefits.
However, we argue that the evaluations of current ILP 
systems 
are insufficient. We demonstrate that the reported results are questionable, especially, in terms of generalization and 
because the datasets are lacking in various dimensions.

The evaluation of rule learning has changed over time.
While the classical rule learning methods often focused on tricky problems in complex domains  \cite{ILPdatasets,Quinlan-ML90:foil} and proved to be effective in practical applications, 
current evaluations can be divided into three categories. 
Some consider very small example problems with usually less than 50 facts and only few rules to be learned \cite{EGre-jair18:learning-explanatory-rules,RoR-NIPS17,minervini2019differentiable}. 
Often, these problems are \emph{completely} defined, in the sense that all facts are classified as either true or false, or there are at least some negative examples given.
Hence, the systems can be thoroughly evaluated based on classical measures such as accuracy.
Other evaluations regard (subsets of) real KGs such as Wikidata\footnote{\url{https://www.wikidata.org/wiki/Wikidata:Main_Page}}
or DBpedia\footnote{\url{https://wiki.dbpedia.org/}}, 
some with millions of facts
\cite{Galarraga+-VLDBJ15:amiep,OWaWa-IJCAI18:scalable-rule-learning,Ho+-ISWC18:guided-by-embedding,RDFRules}. 
Since there are no rules over these KGs, the rule suggestions of the systems are usually evaluated using metrics capturing the precision and coverage of rules (e.g., standard confidence \cite{Galarraga+-VLDBJ15:amiep})
based on the facts contained in the KG.
However, since the KGs are generally incomplete, the quality of the rule suggestions is not fully captured in this way. For instance, \citet{OWaWa-IJCAI18:scalable-rule-learning} present an illustrative example rule, 
$\expred{gender}(X,\exconst{male})\ass\expred{isCEO}(X,Y),\expred{isCompany}(Y)$,
which might well capture the facts in many existing KGs but which is heavily biased and does not extend to the entirety of valid facts beyond them.
Furthermore, we cannot assume that the few considered KGs completely capture the variety of existing domains and especially the rules in them. For example, \citet{Minervini+-NAMPI18:ntp-at-scale} propose rules over WordNet\footnote{\url{https://wordnet.princeton.edu/}} that are of very simple nature -- containing only a small number of the predicates in WordNet and having only a single body atom -- and very different from the ones suggested in 
\cite{Galarraga+-VLDBJ15:amiep} for other~KGs.

Also the evaluation metrics vary, especially considering the intersection between more modern and classic approaches. We will show that most of the standard information retrieval measures used in machine learning are not adequate for a logic context because they neglect important facets like the size of the Herbrand universe (e.g., this may yield a too high accuracy).
Some other measures have been used for neural ILP such as Mean Reciprocal Rank, or precision/recall@K, but they can be applied only in specific cases (i.e. the system outputs weighted/probabilistic rules or a ranking of facts).
Yet, strict logic measures are not perfect either, since they are based on the assumption that the domain is very small and human understandable. For this reason the community needs to consider several metrics and should define new metrics suitable for both worlds.

Recently, synthetic datasets have been proposed, but they are very simple and do not cover all characteristics necessary to evaluate an ILP tool properly: 
1) \citet{Dong+-ICLR19:nlms} provide a first synthetic dataset generator for graph reasoning,
which can produce an arbitrary number of facts regarding five fixed predicates while the rules are hand written.
2) \citet{on_the_fly} argue, in line with us, for 
more diverse datasets for rule learning. However, their generated datasets are still restricted in several dimensions: e.g. small size and 
very simple rules (based on five fixed templates).
There are well-known ILP competitions\footnote{for example: 2016, http://ilp16.doc.ic.ac.uk/competition} in the logic community, but their 
evaluations are usually based only on test facts and not on 
rules. 

There are also benchmarks in the database community which are related in that they cover schemata and rules. However, either their use cases and thus the kinds of rule sets considered are rather restricted (e.g., for schema mapping, there are rules from source to target schema, but the rules do not depend on each other in the sense that one is to be applied after the other \cite{Alexe08-stbench,ibench}), or there is a number of fixed test scenarios \cite{Benedikt17-chasebench}. Further, database data is usually curated, and the benchmarks were developed to evaluate customized algorithms, while our focus is on more arbitrary data, and on learning systems and approximate solutions, needing large amounts of data to learn from (i.e., instead of a few, fixed test sets).

In summary, we claim that the existing 
datasets are not sufficient to cover the possible variety of real data and the rules that could be mined from such arbitrary data. 
However, many existing KGs are large, noisy, heterogeneous, and might embed complex rules. The problem is that we do not know if such embedded rules do not exist or 
if they are just not learned today because of the restrictions of the current rule learners. 
Since it is unclear what sort of complexity is required to model the real world, we opted for an artificial but largely random approach 
that covers different kinds of variety and complexity missing in today's datasets.

In this paper, 
we extend the categorization of rule learning datasets beyond the numbers of constants, predicates, and facts. In particular, we propose to consider rather obvious characteristics like the amount of noise (i.e., wrong or missing facts) and  (in)completeness (i.e., share of consequences of rules present in the data).
 
We present \tool (\textbf{S}ynthetic \textbf{Da}tasets for \textbf{Ru}le Learning), a tool for generating synthetic datasets containing both facts and rules, and for evaluating rule learning systems, that overcomes the above mentioned shortcomings of existing datasets and offers proper evaluation methods. 
\tool is parameterizable in the standard and in the new categories, and thus allows for a more fine-grained analysis of rule learning systems. It also supports this analysis 
by computing classical and more recent metrics, including two new ones that we introduce.
Finally, we evaluate representatives of different types of rule learning systems on our datasets 
demonstrating the necessity of having a diversified portfolio of datasets to help revealing the variety in the capabilities of the systems, and thus also to support and help researchers in developing and optimizing new/existing approaches.
Moreover, our experimental shows in detail the benefits of our two novel metrics, and how they complement the existing performance measures.
%
\tool is available at 
$<$\url{https://github.com/IBM/RuDaS}$>$.

\section{Rule Learning Preliminaries}\label{sec:motivation}

We assume the reader to be familiar with first-order logic (FOL) and its related concepts (e.g., inference, Herbrand models and universes, etc.).
We consider datalog \emph{rules} \cite{datalog}: 
\begin{equation}\label{eq:rule}
 \alpha_0\ass\alpha_1,\dots,\alpha_m .
\end{equation}
of \emph{length} $m\ge1$ where all \emph{atoms} $\alpha_j$, $0\le j\le m$, are of the form $p(t_1,\dots,t_n)$ with a {predicate} $p$ of arity $n\ge1$ and terms $t_k$, $1\le k\le n$.
A \emph{term} is either a constant or a variable. 
$\alpha_0$ is called the \emph{head} and the conjunction $\alpha_1,\dots,\alpha_m$ the \emph{body} of the rule. All variables that occur in the head must occur in the body. A \emph{fact} is an atom not containing variables.

Note that several classical ILP systems also consider more complex function-free Horn rules, which allow for existential quantification in the rule head or negation in the body, but most recent systems focus on datalog rules or restrictions of those \cite{Galarraga+-VLDBJ15:amiep,EGre-jair18:learning-explanatory-rules,RoR-NIPS17}. In particular, reasoning systems for KGs \cite{YaYaCo-NIPS17:neurallp,OWaWa-IJCAI18:scalable-rule-learning}
often consider only binary predicates and \emph{chain rules} 
of the form:
\begin{equation}\label{eq:chainrule}
 p_0(X_1,X_{m+1})\ass p_1(X_1,X_2),\dots,p_{m}(X_{m},X_{m+1}). 
\end{equation}
We define the problem of rule learning in the most general way:
given 
background knowledge in the form of facts, including a set of so-called \emph{positive examples} (vs. \emph{negative} or \emph{counter}-examples), the goal is to learn rules that can be used to infer the positive examples from the background knowledge, based on standard FOL semantics.
As it is common, 
we do not separate the background knowledge into two types of facts but consider a single fact set as input.


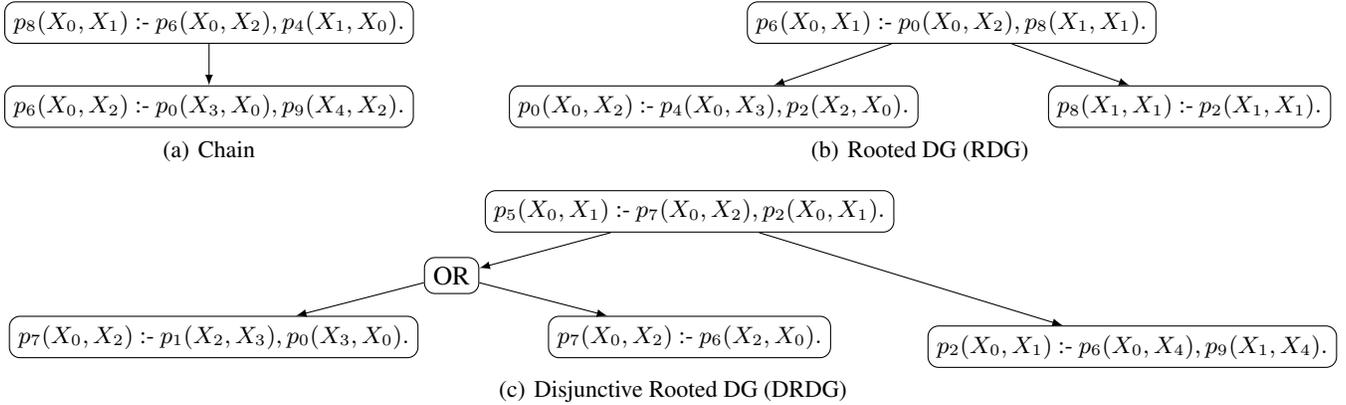
\begin{figure*}[h!]
\centering 
\begin{adjustbox}{minipage=\linewidth}
    \centering
    \subfigure[Chain]{
        \centering
        \begin{tikzpicture}[edge from parent/.style={draw,-latex},sibling distance=10em,every node/.style = {shape=rectangle, rounded corners, draw, align=center, top color=white,transform shape, font={\fontsize{9pt}{0}}}]]]
            \node {$p_8(X_0,X_1) \ass p_6(X_0,X_2),p_4(X_1,X_0).$}
                child { node[above=0.1cm] {$p_6(X_0,X_2) \ass p_0(X_3,X_0),p_9(X_4,X_2).$}}
            ;
        \end{tikzpicture}} 
            \hspace{1cm}
  \subfigure[Rooted DG (RDG)]{
        \centering
        \begin{tikzpicture}[edge from parent/.style={draw,-latex},sibling distance=18em,
        every node/.style = {shape=rectangle, rounded corners,
        draw, align=center,
        top color=white,transform shape},font={\fontsize{9pt}{0}}]]
            \node {$p_6(X_0,X_1) \ass p_0(X_0,X_2),p_8(X_1,X_1).$}
                child { node[above=0.1cm] {$p_0(X_0,X_2) \ass p_4(X_0,X_3),p_2(X_2,X_0).$}}
                child { node[above=0.1cm] {$p_8(X_1,X_1) \ass p_2(X_1,X_1).$}}
            ;
        \end{tikzpicture}
    }
    \\
    \subfigure[Disjunctive Rooted DG (DRDG)]{
        \centering 
        \begin{tikzpicture}[edge from parent/.style={draw,-latex}, sibling distance=18em,
        every node/.style = {shape=rectangle, rounded corners,
        draw, align=center,
        top color=white,transform shape},font={\fontsize{9pt}{0}}]
        ]
            \node {$p_5(X_0,X_1) \ass  p_7(X_0,X_2),p_2(X_0,X_1). $}
            child{ node[above=0.4cm]{OR}
                child { node[above=0.4] {$p_7(X_0,X_2) \ass p_1(X_2,X_3),p_0(X_3,X_0).$}}
                child { node[above=0.4] {$p_7(X_0,X_2) \ass p_6(X_2,X_0).$}}}
            child{node[above=-0.3,right] {$p_2(X_0,X_1) \ass p_6(X_0,X_4),p_9(X_1,X_4).$}};
        \end{tikzpicture}
    }%
    \end{adjustbox}
    \caption{Example rule structure generated for the different categories  with size S and depth 2.
    }
    \label{fig:datasets}
\end{figure*}

\subsection*{Approaches to Rule Learning} 

Classical ILP systems such as FOIL \cite{Quinlan-ML90:foil} and Progol \cite{Muggleton-NGC95:progol} usually apply exhaustive algorithms to mine rules for the given data and either require false facts as
counter-examples or assume a closed world (for an overview of classical ILP systems see Table~2 in \cite{SteGaHo-RR18:overview}).
The \emph{closed-world assumption} (CWA) (vs. {\it open world assumption} or OWA) states that all facts 
that are not explicitly given as true are assumed to be false.

Today, however, knowledge graphs with their often incomplete, noisy, heterogeneous, and, especially, large amounts of data raise 
new problems and require new solutions.
For instance, real data most often only partially satisfies the CWA
and does not contain counter-examples. Moreover, in an open world, absent facts cannot be considered as counter-examples either, since they are not regarded as false.
Therefore, successor systems, with AMIE+ \cite{Galarraga+-VLDBJ15:amiep} and RDF2Rules \cite{WangLi-CoRR15:rdf2rules} as the most prominent representatives, 
assume the data to be only partially complete and
focus on rule learning in the sense of mining patterns that occur frequently in the data.
Furthermore, they implement advanced optimization approaches that make them applicable in wider scenarios. 
In this way, they address already many of the issues that arise with today's knowledge graphs, still maintaining their processing exhaustive.

Recently, {neural rule learning} approaches have been proposed:  \citet{YaYaCo-NIPS17:neurallp,RoR-NIPS17,EGre-jair18:learning-explanatory-rules,Minervini+-NAMPI18:ntp-at-scale,OWaWa-IJCAI18:scalable-rule-learning,Campero+-corr18}. 
These methodologies seem a promising alternative 
considering that deep learning copes with vast amounts of noisy and heterogeneous data.
The proposed solutions consider vector or matrix embeddings of symbols, facts and/or rules, and model inference using differentiable operations such as vector addition and matrix composition. However, they are still premature: they only learn certain kinds of rules or lack scalability (e.g., searching the entire rule space) and hence cannot compete with established rule mining systems such as AMIE+ yet, as shown in \cite{OWaWa-IJCAI18:scalable-rule-learning}, for example. 

\section{\tool Datasets}\label{sec:description}

\tool contains an easy-to-use generator 
for ILP datasets that generates datasets that vary in many dimensions and is highly parameterizable. While existing datasets are missing more detailed specifications but are described only in terms of size and number of different constants and predicates, we propose a much more detailed set of metrics (parameters in the generator) which can serve as a general classification scheme for ILP datasets and support evaluations.
In this section, we give details about these metrics, and thus on the possible shapes of \tool datasets. 
Each dataset contains the rules and the facts in files in standard Prolog format (using the syntax of Rule~(\ref{eq:rule})). 
We also describe example datasets we generated, which can be found in our repository. 
Observe that our datasets are based on rules, as required for ILP, but can be applied for evaluating the closely related and very popular link prediction systems as well;  specifically, auxiliary information provided with the datasets contains the missing consequences which can be used as test facts in that context.


\textbf{Symbols.} 
Our datasets are domain independent, which means that we consider synthetic names $p_i$ for 
{predicates}, $c_i$ for {constants}, and $X_i$ for {variables} with $i\ge0$. 
While the kinds and numbers of the symbols used is random, it can be controlled by setting the following generator parameters:
\begin{itemize}
\item number of constants and predicates
\item min/max arity of predicates
\end{itemize}
Observe that these numbers influence the variability and number of generated rules and facts.



%
\textbf{Rules.} 
\tool datasets contain 
datalog {rules} (see Section~\ref{sec:motivation}) of variable structure. The generation is largely at random in terms of which predicates, variables, and constants appear in the rules; that is, in the structure of every single rule. We only require the head to contain some variable.
%
%

To classify a set of rules, we propose four categories depending on the dependencies between rules: \emph{Chain}, \emph{Rooted Directed Graph (DG)}, \emph{Disjunctive Rooted DG}, and \emph{Mixed}. 
Figure~\ref{fig:datasets} shows a generated rule set for each category. 
The dependencies between the rules are represented as edges in a directed graph 
where the rules are the nodes. That is, an incoming edge shows that the facts inferred by the child node's rule might be used, during inference 
with the rule at the parent node.
The 
node at the top
is called the \emph{root}. In the following, we use (rule) graph and DG interchangeably. 

\textit{Category Chain.} Each rule, except the one at the root, infers facts relevant for exactly one other rule (i.e., every node has at most one parent node) and, for each rule, there is at most one such other rule which might infer facts relevant for the rule (i.e., every node has at most one child node). However, recursive rules (where the predicate in the head occurs also in the body) represent an exception, they are relevant for themselves and for one other rule (i.e., the graph has a small loop at each node representing a recursive rule).

\textit{Category Rooted DG (RDG).} It generalizes category Chain in that every rule can be relevant for several others (i.e., each node can have multiple parent nodes).
Furthermore, for each rule, there may be several other rules which might infer facts relevant for the rule (i.e., a node may have several child nodes); 
and at least one such case exists.
But, for each predicate occurring in the body of the former rule, there must be at most one other rule with this predicate in the head; 
that is, there are no alternative rules to derive facts relevant for a rule w.r.t.\ a specific body atom. 

\textit{Category Disjunctive Rooted DG (DRDG).} It generalizes category RDG by allowing for the latter alternative rules (represented as children of an ``OR'' node); and at least one such case exists.

\textit{Category Mixed.} A rule graph that contains connected components of different of the above categories. 


\begin{table*}[t!]
    \centering
    \small
    \begin{tabular}{lcccccccccccccccccccc}
    \toprule
     \multirow{2}{*}{\#} & \multirow{2}{*}{Rule type}   & \multirow{2}{*}{Size} & \multirow{2}{*}{Depth} &&  \multicolumn{3}{c}{\#Rules} &&\multicolumn{3}{c}{\#Facts} 
      &&\multicolumn{3}{c}{\#Pred} && \multicolumn{3}{c}{\#Const} \\ 
      \cmidrule{6-8}
      \cmidrule{10-12}
      \cmidrule{14-16}
      \cmidrule{18-20}
      &&& 
      && min&avg&max
      & &min&avg&max
      & &min&avg&max
      & &min&avg&max
      \\
     \midrule
     10 &CHAIN  &   XS & 2 &&2&2&2 && 51&74&95 && 5&7&9 && 31&47&71  \\
          10 &CHAIN  &   XS & 3 &&3&3&3& & 49&70&97 && 7&8&9 && 31&43&64 \\
     10 & CHAIN  & S & 2  && 2&2&2 && 168&447&908 && 9&10&11 && 97&259&460\\
      10 &CHAIN  & S &  3 && 3&3&3 && 120&508&958 & &8&10&11 && 52&230&374\\
     \midrule
    22 & RDG  &  XS & 2  && 3&3&3 && 49&84&122 && 6&9&11 && 28&50&84 \\ 
     12 & RDG  &  XS & 3  && 4&5&6 && 56&104&172 && 8&10&11 && 41&55&75 \\
     22 &  RDG  & S & 2 && 3&3&3 && 200&646&1065 && 6&11&11 && 71&370&648   \\
     22 &RDG  &  S & 3 && 4&5&7 && 280&613&1107 && 10&11&11 && 149&297&612   \\
     \midrule
     22 &DRDG &  XS & 2  && 3&4&5 && 60&100&181 && 6&9&11 && 29&55&82    \\
     12 &DRDG &  XS &3  && 4&7&11 && 58&144&573 && 8&10&11 && 34&58&89   \\
     22 &DRDG &  S &2 && 3&4&5 && 149&564&1027 && 10&11&11 && 88&327&621   \\
     22 &DRDG &  S &3 && 4&7&12 && 111&540&1126 && 10&11&11 && 70&284&680   \\
     \bottomrule
    \end{tabular}
    \caption{Overview of our generated datasets, altogether 196;
    column \# is the count of datasets described in the corresponding row.
    All other numbers are averages.
    For Chain, we have $\nowa=0.3$, $\nnoisemiss=0.2$, and $\nnoiseadd=0.1$.
    For RDG and DRDG:
    $\nowa\in\{0.2,0.3,0.4\}$, $\nnoisemiss\in\{0.15,0.2,0.3\}$, and $\nnoiseadd\in\{0.1,0.2,0.3\}$.
    Note that the size bounds of our fact sets are not strict, some sizes are slightly larger than expected (e.g., 1065 for size S) because our initial generation needs to take into account that some facts, e.g., consequences, may be removed thereafter. 
    }
    \label{tab:datasets}
\end{table*}

Figure~\ref{fig:datasets} illustrates the differences between the categories. In (a), for each rule, there is at most one child node with a rule relevant for its derivations.
In (b), there might be multiple children, but each child node contains a different head predicate. In (c), the latter does not hold anymore; for given facts, there may be various derivations. 

The numbers and categories of connected components are selected randomly by default.
The shape of \tool rule sets can be influenced with the following parameters though: 

\begin{itemize}
\item number and maximal length of rules
\item category of connected components (i.e., one of the above)
\item min/max number of connected components
\item maximal depth of rule graphs (i.e., number of rules nodes in the maximum of the shortest paths between root and leaves)
\end{itemize}

\textbf{Facts.} 
The main advantage of the \tool datasets, the availability of the rules, allows for classifying the facts as well. More specifically, facts can be \emph{(ir)relevant} for inference, depending on if their predicates do (not) occur in a rule body, and they may be consequences of inferences.
Such a classification of facts is impossible for all the existing datasets that do not contain rules, but allows for a better evaluation of the rule learners' capabilities (see Section~\ref{sec:experiments}).  

\tool fact sets vary in the following parameters:
\begin{itemize}
\item dataset size: XS, S, M, L, XL
\item open-world degree \nowa $\in [0,1]$ 
\item amount of noise in the data \nnoiseadd, \nnoisemiss $\in [0,1]$ 
\end{itemize}
An XS dataset contains about 50-100 facts, 
an S dataset about 101-1,000, 
an M dataset about 1,001-10,000,
an L dataset about 10,001-100,000,
and an XL dataset about 100,001-500,000. 
For larger sizes, we suggest meaningful abbreviations in the form of X2L for XXL etc., which allow for extension while being short and easy to understand.
Since the main purpose of \tool is allowing the analysis of the rules learned (vs.\ scalability), we have however not considered such larger datasets so far. 
The open-world degree \nowa specifies how many of the consequences from an initial set of relevant facts, called \emph{\support} facts, are missing in the dataset (see Section~\ref{sec:generation} for a detailed description of the generation process).
By noise, we mean facts that are not helpful in learning the rules either because they are not relevant for deriving the positive examples (\nnoiseadd) 
or because they are relevant but missing (\nnoisemiss).

\subsection*{Example Datasets: \tool-v0}\label{sec:generatedDatasets}

For demonstration purposes, we generated \tool-v0, a set of datasets which are available to the community (in our repository), and which we also used in our experiments (see Section~\ref{sec:experiments}).
Table~\ref{tab:datasets}
shows statistics about them.
The datasets model different possible scenarios, and mainly vary in the structures and sizes of the rule sets and in the sorts and quantities of facts. 
\tool-v0 contains 40 Chain, 78 RDG, and 78 DRDG datasets, of sizes XS and S, and of depths 2 and 3, all evenly distributed.
%
Note that each of the rules sets in \tool-v0 consists of exactly one connected component, and that we did not generate rule sets of category Mixed; Mixed datasets with connected components of possibly different categories can be easily created by combining our generated datasets.  
Further, we constrained both the maximal rule length and arity of atoms to two 
because several existing rule learning systems require that.

All the datasets were generated such that they are missing 20-40\% of all consequences,  15-30\% of the original support facts, and contain 10-30\% facts that are irrelevant for the derivation of positive examples.
Since real datasets may strongly vary in the numbers of missing consequences and noise and, in particular, since these numbers are generally unknown, we chose factors seeming reasonable to us. Also note that 
there is information regarding
the accuracy of real fact sets such as YAGO\footnote{\url{https://github.com/yago-naga/yago3}} (95\%) and NELL\footnote{\url{http://rtw.ml.cmu.edu/rtw/overview}} (87\%), that measures the amount of data correctly extracted from the Web etc. and hence corresponds to  $1-\nnoiseadd$ in our setting. Thus, our choices in this regard thus seem to be realistic.
%

We hence simulated an open-world setting and incorporated noise. While we consider this to be the most realistic training or evaluation scenario, 
specific rule learning capabilities might be better evaluated
in more artificial settings with either consequences or noise missing. For this reason, every dataset mentioned in the table additionally includes files containing the incomplete set of facts without noise (i.e., $\nowa$ as in the table; $\nnoiseadd=0$; $\nnoisemiss=0$) and the 
complete fact set (i.e., $\nowa=0$), with and without noise.

\section{Dataset Generation}\label{sec:generation}

In this section, we describe the generation process of the rules and facts in detail, assuming the generator parameters (also \emph{configuration}) listed in Section~\ref{sec:description} to be set.


\textbf{Preprocessing.}
As already mentioned, most parameters are determined randomly in a preprocessing step if they are not fixed in the configuration, such as the symbols that will be used, the numbers of DGs to be generated, and their depths.
However, all random selections are within the bounds given in the configuration under consideration; for instance,
we ensure that the symbols chosen suffice to generate rule graphs and fact sets of selected size and that at least one graph is of the given maximal depth.

\textbf{Rule generation.}
According to the rule set category specified and graph depths determined, rules (nodes in the graphs) of form (\ref{eq:rule}) are generated top down breadth first, for each of the rule graphs to be constructed. 
The generation is largely at random, that is, w.r.t.
the number of child nodes of a node and which body atom they relate to; 
the number of atoms in a rule; and the predicates within the latter, including the choice of the \emph{target predicate} (i.e., the predicate in the head of the root) in the very first step. 
\tool also offers the option that all graphs have the same target predicate.
%
To allow for more derivations, we currently only consider variables as terms in head atoms; the choice of the remaining terms is based on probabilities as described in the following. 
Given the atoms to be considered (in terms of their number and predicates) and an arbitrary choice of head variables, we first determine a position for each of the latter in the former. Then we populate the other positions one after the other: a head variable is chosen with probability $p_h=\frac{1}{5}$; for one of the variables introduced so far, we have probability $p_v=(1-p_h) * \frac{3}{4}$; for a constant, $p_c=(1-p_h) * (1-p_v) * \frac{1}{10}$; and, for a fresh variable, $p_f=(1-p_h) * (1-p_v) * (1-p_c)$. While this conditional scheme might seem rather 
complex, we found that it works best in terms of the variety it yields; also, these probabilities can be changed easily.


\textbf{Fact generation.}
The fact generation is done in three phases: we first construct a set \db of relevant facts in a closed-world setting, consisting of {\support facts} \sfacts and their {consequences} \cfacts, and then adapt it according to \nowa and \nnoise. 

As it is the (natural) idea, we generate facts by instantiating the rule graphs multiple times,
based on the assumption that rule learning systems need positive examples for a rule to learn that rule, and stop the generation when the requested number of facts has been generated. We actually stop later because we need to account for the fact that we subsequently will delete some of them 
according to \nowa. 
More specifically, we continuously iterate over all rule graphs, for each, select an arbitrary but fresh variable assignment $\sigma$, and then iterate over the graph nodes as described in the following, in a bottom-up way. 
%
First, we consider each leaf $n$ and corresponding rule of form (\ref{eq:rule}) 
and generate support facts $\sigma(\alpha_1),\dots,\sigma(\alpha_m)$.
Then, we infer the consequences based on the rules and all facts generated so far. 
For every node $n$ on the next level and corresponding rule of form (\ref{eq:rule}), we only generate those of the facts $\sigma(\alpha_1),\dots,\sigma(\alpha_m)$ as support facts which are not among the consequences inferred previously. We then again apply inference, possibly obtaining new consequences, and continue iterating over all nodes in the graph in this way.
We further diversify the process based on two integer parameters, \ndag and \nnskip: in every \ndag-th iteration the graph is instantiated exactly in the way  described; in the other iterations, we skip the instantiation of a node with probability 1/\nnskip and, in the case of DR-DGs, only instantiate a single branch below disjunctive nodes. 
We implemented this diversification to have more variability in the supports facts, avoiding to have only complete paths from the leaves to the root.

In the open-world setting, we subsequently construct a set \dbowa by randomly deleting consequences from \db according to the open-world degree given: 
assuming $\tfacts\subseteq\cfacts$ to be the set of \emph{target facts} (i.e., consequences containing the target predicate), 
we remove $\nowa\%$ from $\cfacts\setminus\tfacts$, and similarly $\nowa\%$ from \tfacts.
In this way, we ensure that the open-world degree is reflected in the target facts. Though, there is the option to have it more arbitrary by removing $\nowa\%$ from $\cfacts$ instead of splitting the deletion into two parts.

The noise generation is split similarly. 
Specifically, we construct a set \dbowan based on \dbowa by arbitrarily removing $\nnoisemiss\%$ from \sfacts, and by adding arbitrary fresh facts that are neither in \cfacts (i.e., we do not add facts we have removed in the previous step) nor contain the target predicate
such that $\dbowan\setminus\tfacts$ contains $\nnoiseadd\%$ of noise.
In addition, we add arbitrary fresh facts on the target predicate that are not in \tfacts already 
such that subset of \dbowan on that predicate finally contains $\nnoiseadd\%$ of noise.

\textbf{Output.}
The dataset generation produces:
the rules; a \emph{training} set (\dbowan), which is of the requested size, and fulfills \nowa, \nnoiseadd, and \nnoisemiss; and custom fact sets $\sfacts'$ and $\cfacts'$ for our evaluation tools generated in the same way as \sfacts and \cfacts. 
For further experiments, \tool 
also outputs 
\db, \dbn (an adaptation of \db which contains noise but all of \cfacts), \dbowa, \sfacts, and~\cfacts (see also the end of Section~\ref{sec:generatedDatasets}).

\section{Evaluation Tools}\label{sec:evaluation}

\tool contains also an evaluator (written in Python) that is able to compare the original rules of a dataset to the ones produced by a rule learning system. 
In the following, we give details about its evaluation approach and about the different measures we consider.

We focus on three logic(-inspired) distances and four standard information retrieval measures that are relevant to our goal of capturing rule learning performance: 
1) Herbrand distance, the traditional {distance between Herbrand models}; two normalized versions of the Herbrand distance 2) Herbrand accuracy (H-accuracy) and 3) Herbrand score (H-score), a new metric we propose in this paper; 4) accuracy 5) precision;  6) recall; and 7) F1-score. 


Our test fact sets (both facts and consequences) in the evaluation do not contain noise and all the consequences can be recovered by the original rules applied over the given facts. In line with that, we focused on measures that maintain the closed-world assumption, and did not include in \tool measures that focus on the open-world aspect for the evaluation (i.e., {\it PCA} in \cite{Galarraga+-VLDBJ15:amiep}). 
Although, as it is explained in Section~\ref{sec:ml_measures}, F1-score is the best suit metric in \tool to deal with an open-world evaluation. 

In what follows, $I(R,F)$ denotes the set of facts inferred by grounding the rules $R$ over the support facts $F$ excluding the facts in $F$.
We denote an original rule set by $\mathcal{R}$, a learned one by $\mathcal{R}'$, and  support facts by $\mathcal{F}$.
Our evaluation is performed comparing two sets: 
1) $I(\mathcal{R}',\mathcal{F})$ obtained by the application of the induced rules $\mathcal{R}'$ to the fact sets $\mathcal{F}=\sfacts'$ ($\sfacts'$ and $\cfacts'$ described in Section~\ref{sec:generation} - Output) using a forward-chaining engine (written in python and available in our tool);
2) $\cfacts'$ that corresponds to $I(\mathcal{R},\mathcal{F})$: the result of the application of the original rules $\mathcal{R}$ to the fact set $\mathcal{F}=\sfacts'$.

\subsection{Logic Measures}
The \emph{Herbrand distance} $h_d$ between two logic programs (sets of rules), defined over the same set of constants and predicates, is defined as the number of facts that differ between the two minimal Herbrand models of the two programs: 
$$
h_d(\mathcal{R},\mathcal{R}',\mathcal{F}):= 
|~
[I(\mathcal{R},\mathcal{F}) \cup I(\mathcal{R}',\mathcal{F})]
\setminus 
[I(\mathcal{R},\mathcal{F}) \cap I(\mathcal{R}',\mathcal{F})]
~ | ~.
$$

The \emph{standard confidence} $s_c$ \cite{Galarraga+-VLDBJ15:amiep} 
is the fraction of correctly inferred facts w.r.t. all facts that can be inferred by the learned rules capturing their precision: 
$$s_c(\mathcal{R},\mathcal{R}',\mathcal{F})
:= \frac{|I(\mathcal{R},\mathcal{F})\cap I(\mathcal{R}',\mathcal{F})|}{|I(\mathcal{R}',\mathcal{F})|} $$
In our closed-world setting, this corresponds to the precision of a model, since it is easy to see that  $ |I(\mathcal{R},\mathcal{F})\cap I(\mathcal{R}',\mathcal{F})|$  are the number of true positive examples and $|I(\mathcal{R}',\mathcal{F})|$ corresponds to the union of true and false positive examples.
The {\it Herbrant accuracy}  $h_r$ 
corresponds to the Herbrand distance normalized on the Herbrand universe: $h_r(\mathcal{R},\mathcal{R}',\mathcal{F}):= 1- \frac{h_d}{u}$, where $u$ is the size of the Herbrand universe defined by the original program.
We introduce a new metric, the \emph{Herbrand score} (H-score) 
defined as:
\begin{align*}
\text{H-score}(\mathcal{R},\mathcal{R}',\mathcal{F}):&= \frac{|I(\mathcal{R},\mathcal{F})\cap I(\mathcal{R}',\mathcal{F})|}{|I(\mathcal{R},\mathcal{F})\cup I(\mathcal{R}',\mathcal{F})|} \\
&= 1- \frac{h_d(\mathcal{R},\mathcal{R}',\mathcal{F})}{|I(\mathcal{R},\mathcal{F})\cup I(\mathcal{R}',\mathcal{F})|} 
\end{align*}

H-score provides an advantage over the other metrics since it captures both how many correct facts a set of rules produces and also its completeness (how many of the facts inferred by the original rules $\mathcal{R}$ were correctly discovered), while the other measures consider these points only partially.

Note that Herbrand accuracy is not a significant measure if $\mathcal{F}$ or the Herbrand universe is large, because, in these cases, it will be very high (close to $1$) disregarding the quality of the rules. This happens because all the facts in $\mathcal{F}$ are considered correct predictions, as well as the facts in the Herbrand universe that neither appear in $I(\mathcal{R},\mathcal{F})$ nor in $I(\mathcal{R}',\mathcal{F})$.

\subsection{Information Retrieval Measures}\label{sec:ml_measures}
We adapted the main measures used in the machine learning evaluations to our context.
We define: the sets of true positive examples (TP) as the cardinality  $|I(\mathcal{R},\mathcal{F}) \cap I(\mathcal{R}',\mathcal{F})|$, the set of false positive examples (FP) as the cardinality $|I(\mathcal{R},\mathcal{F}) \setminus I(\mathcal{R}',\mathcal{F})|$; the set of false negative examples (FN) as the cardinality $|I(\mathcal{R}',\mathcal{F}) \setminus I(\mathcal{R},\mathcal{F})|$; and the set of true negative examples (TN) as the cardinality of the difference between the Herbrand universe and the union  $I(\mathcal{R},\mathcal{F}) \cup I(\mathcal{R}',\mathcal{F})$.
Given these four definitions, accuracy, precision, recall, F1-score etc. can be defined as usual \cite{Russell:2009:AIM:1671238}.

Note that the accuracy measure is not a significant measure if $\mathcal{F}$ or the Herbrand universe is large, for the same reason reported for Herbrand accuracy above. 
Moreover, F1-score is similar to H-score, with the difference that F1-score gives more priority to the TP examples. We believe that giving uniform priority to FN, TP, and FP is more reasonable in the context of logic; this is in line with standard logic measures like $h_d$. However, F1-score better suits (compared to H-score) open-world settings where some of the consequences could be missing, and thus count as FP (despite being correct). For this reason F1-score would give a better estimate of the quality of the induced rules since it focuses more on the TP examples and give less priority to the generated FP examples.

We observe that, if $I(\mathcal{R},\mathcal{F}) = I(\mathcal{R}',\mathcal{F})$, then H-score is equal to precision 
and both are equal to $1$; and, if $I(\mathcal{R},\mathcal{F})$ and $I(\mathcal{R}',\mathcal{F})$ are disjoint, then both are $0$. Moreover the two measures coincide if $I(\mathcal{R},\mathcal{F}) \subseteq I(\mathcal{R}',\mathcal{F})$. 
The main difference between the two measures is highlighted in the case where $I(\mathcal{R}',\mathcal{F}) \subseteq I(\mathcal{R},\mathcal{F})$. 
Then, precision $=1$ but H-score is $<1$. This property is intentional for our new metric (H-score) because we want to have H-score $1$ only if the predicted facts are exactly those
produced by the original rules while precision is $1$ as soon as all predicted facts are correct.

\subsection{Rule-based Measures}
Several distance metrics between two sets of logic rules have been defined in the literature \cite{dist_1,dist_2,dist_3,dist_4,dist_5}. 
However, 
most of them strongly rely on the parse structure of the formulas and hence are more suitable for more expressive logics. 

For this reason, we propose a new metric, the \emph{Rule-score (R-score)}, which is tailored to rules, 
in that it calculates a distance $d_R$ between the rules that have the same head predicate and, for the latter, computes the pairwise distances $d_A$ between the two rules' atoms. Note that, below, we consider datalog rules without constants and with only binary predicates for simplicity, 
but the definitions can be easily extended to non-binary atoms. 
An idea of how to deal with constants in addition to variables is suggested in \cite{dist_2}, for example.


Our distance $d_A$ between two non-ground, binary atoms takes into account a specific mapping $\omega \in \Omega$ between their variables so that we can later lift it to rule level;
$\Omega$ is the set of all possible 
\emph{variable re-namings} between two sets of variables names (in our case from two rules). 
Given two atoms $a_1=p(X_1,X_2)$ and $a_2=q(Y_1,Y_2)$ and such a mapping $\omega$,  $d_A(a_1,a_2,\omega)=1$ if their predicates differ, $d_A(a_1,a_2,\omega)=0 $ if 
$\omega(X_1)=Y_1$ and $\omega(X_2)=Y_2$
, and otherwise:
$$
 d_A(a_1,a_2,\omega) = 
 \frac{1}{4} \sum\limits_{i =1}^2 
 \mathds{1}^c(\omega(X_i)=Y_i)
 ~.$$
where the \emph{complement indicator function} $\mathds{1}^c(e)$ of an event $e$ is equal to $0$ if the event is satisfied and $1$ otherwise.

Note that $d_A$ is based upon the main, and most used, distance metric between two \emph{ground} atoms, the \emph{Nienhuys-Cheng distance} \cite{Nienhuys_Cheng_distance}. In fact, it only differs from the latter in that it is defined for non-ground atoms by considering a variable re-naming.

\begin{example}\label{ex:1}
Consider two rules $r_1$ and $r_2$ as follows:
\begin{align*}
r_1: &\hspace{0.3cm} p_1(A,B) \ass p_2(A,A), p_3(B,B), p_4(A,B). \\
r_2: &\hspace{0.3cm} p_1(X,X) \ass p_2(Y,X), p_2(X,X).
\end{align*}
The re-namings 
$\omega_1=\{A:X,B:Y\}$ and $\omega_1=\{A:Y,B:X\}$ yield:
\begin{itemize}
\item[1)] $r_1$ with $\omega_1$: $\hspace{0.2cm} p_1(X,Y) \ass p_2(X,X), p_3(Y,Y), p_4(X,Y).$
\item[2)] $r_1$ with $\omega_2$: $\hspace{0.2cm} p_1(Y,X) \ass p_2(Y,Y), p_3(X,X), p_4(Y,X).$
\end{itemize}
Let $h_i$ and $h_2$ denote the head atoms of $r_1$ and $r_2$, respectively. Then, we get 
$d_A(h_1,h_2,\omega_1)
=0.25$ and $d_A(h_1,h_2,\omega_2)
=0.25$.
\end{example}

Intuitively, our distance $d_R$ between two rules $r_1$ and $r_2$ considers matches between the rules, where a match consists of a pairing between their atoms together with a variable renaming, and takes the best match as distance (averaged over the number of atoms).
More specifically, such a \emph{pairing} for $r_1$ and $r_2$ is a set of pairs such that the first component is a body atom from $r_1$ and the second from $r_2$. It contains $\max(|b(r_1)|,|b(r_2)|)$ pairs ($|b(r_i)|$ denotes the number of atoms in the body of $r_i$). 
To represent a match, we require the atoms to have the same predicate and additionally allow for an empty placeholder atom (denoted by $-$), extending $d_A$ such that it has maximal distance $1$ to any other atom. Note that the placeholder also accounts for the case that the rules are of different length. 

\begin{example}
For the rules of Example~\ref{ex:1}, there are two pairings: 
\begin{itemize}
\item[a)]$c_1=\{~(p_2(A,A),p_2(Y,X)),~(p_3(B,B),-),~(p_4(A,B),-)~\}$   
\item[b)]$c_2=\{~(p_2(A,A),p_2(X,X)),~(p_3(B,B),-),~(p_4(A,B),-)~\}$
\end{itemize}
\end{example}

Now, we define the distance $d_R$ between rules $r_1$ and $r_2$
as:
$$
d_R(r_1,r_2) = \frac{1}{n_a}\min_{\omega \in \Omega}\Big( d_A(h_1,h_2,\omega) + \min_{c \in \mathcal{C}} \sum_{(a_1,a_2) \in c} d_A(a_1, a_2, \omega) \Big)
$$
where
$n_a =\max(|b(r_1)|,|b(r_2)|)+1$ is the number of atoms of the rule with more atoms;
$h_i$ is the head atom of rule $r_i$;
$\Omega$ is the set of all possible 
variable re-namings between the variable names in the two rules;
$\mathcal{C}$ is the set of all possible pairings for $r_1$ and $r_2$. 

\begin{example}
For the re-namings $\omega_1$ and $\omega_2$ and pairings $c_1$ and $c_2$ from the previous examples, $d_R$ represents the minimum of $0.25+ \sum_{(a_1,a_2) \in c_i} d_A(a_1, a_2, \omega_j)$, with the sums:
\begin{itemize}
\item[1a)] $0.25 + 1 +1 = 2.5$ \hspace{0.3cm}2a)
\hspace{0.03cm}$0.25 + 1 +1 = 2.5$
\item[1b)] $0 + 1 +1 = 2.25$ \hspace{0.54cm}2b)
\hspace{0.03cm}$0.5 + 1 +1 = 2.75$
\end{itemize}
Given that $n_a=4$, we obtain $d_R(r_1,r_2) = \frac{2.25}{4} = 0.5625$.
\end{example}

Finally, our novel metric \emph{Rule-score (R-score)} for two logic programs, the original program $\mathcal{R}$ and the induced program $\mathcal{R}'$ (rules of form~\eqref{eq:rule}), is defined as:
$$
\text{R-score}(\mathcal{R},\mathcal{R}') = 
1 - \frac{1}{|\mathcal{R}|} 
\Big( \sum_{r_1 \in \mathcal{\mathcal{R}}} 
~ ~ ~ ~ ~ 
\min_{r_2 \in \mathcal{R}'[hp(r_1)]} d_R(r_1,r_2) \Big)
$$
where the function $hp(r)$ corresponds to the head predicate of a given rule $r$; 
$\mathcal{R}[p]$ denotes the rules in $\mathcal{R}$ with head predicate $p$, and $|\mathcal{R}|$ denotes the number of rules in $\mathcal{R}$.

\section{Experiments}\label{sec:experiments}

The goal of our experiments is to demonstrate the necessity of having a diversified portfolio of datasets for the evaluation of a rule learning system. The existing datasets are not diverse enough to provide a comprehensive evaluation of ILP methods (e.g., often fall into category Chain). 
Note that the purpose of our experiments is not to provide an exhaustive analysis of existing ILP systems but to show that this kind of analysis should be done, by pointing out that important aspects of rule learning data that have been ignored so far impact system performance. 

To this purpose we chose a set of representative ILP systems (described below) that covers the different methodologies adopted by researchers during the years to approach the problem of rule learning.

We compared the following systems
(configuration details in the appendix): 
1) FOIL \cite{Quinlan-ML90:foil}, a traditional ILP system; 2) AMIE+ \cite{Galarraga+-VLDBJ15:amiep}, a rule mining system; 3) Neural-LP \cite{YaYaCo-NIPS17:neurallp}; and 4) NTP \cite{RoR-NIPS17}. The latter are both neural approaches.
AMIE+, Neural-LP, and NTP output confidence scores for the learned rules.
We therefore filtered their output using a system-specific threshold, obtained using grid search over all datasets.
Further, to not disadvantage Neural-LP and NTP, which use auxiliary predicates, we ignored the facts produced on these predicates in the computation of the result metrics. It is important to notice that NTP requires additional information in the form of rule templates, obviously representing an advantage. 

We evaluated the systems on the datasets described in Section~\ref{sec:generatedDatasets} in four main experiments to understand, respectively, the variety of the performance metrics, and the impact of missing consequences, noise, rule dependencies, and dataset size.
%
We set a time limit of 24h (both for  system executions and  evaluations\footnote{If a system learns very many (usually wrong) rules, the computation of measures based on a closure may become unfeasible.}) to make the evaluation feasible since some computations took days. However, we did not penalize the instances that exceeded the limit since we are interested in the rules that can be learned overall. 
Generally, the runtimes varied greatly over the datasets, were often surprisingly long (AMIE, NTP can take hours; only Neural-LP usually terminated within seconds), and provide not much insight.

\subsection{Overall Results in Different Metrics}\label{sec:exp1}

\begin{figure}
\centering
\begin{tikzpicture}
\begin{axis}[width=9cm,height=6cm,legend style={at={(0.41,0.98)},
		anchor=north,legend columns=3}, xmin=-0.3,xmax=3.3,ymin=0, ymax=0.4, ymajorgrids,every axis plot/.append style={very thick}, xtick=data,xticklabels={FOIL, NTP ,Neural-LP,AMIE+}]
\addplot+[black,mark=,solid] 
coordinates
{		
(0,0.2151)
(1,0.1125)
(2,0.1621)
(3,0.3031)
};
\addplot+[red,mark=,solid] 
coordinates
{		
(0,0.2728)
(1,0.1811)
(2,0.1906)
(3,0.3350)
};
\addplot+[blue,mark=,solid] 
coordinates
{		
(0,0.1520)
(1,0.0728)
(2,0.1025)
(3,0.2321)
};
\legend{\small F1-score, \small R-score, \small H-score}
\end{axis}
\end{tikzpicture}
\caption{Table~\ref{tab:result_1} results visualized showing  similarity of metrics.
}
\label{fig:measures}
\end{figure}
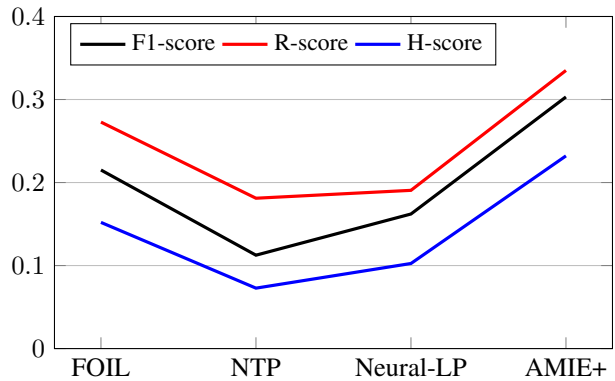

In this experiment, we regard the overall results, reported in Table \ref{tab:result_1}, in terms of the metrics introduced in Section~\ref{sec:evaluation}.
As expected, the results for F1-score and Herbrand score are very similar, the only difference is that F1-score is a more ``optimistic'' measure, giving advantage to the methods with a higher number of true positive examples.
The results for R-score are in line with these metrics and, although translated, follow especially close the trend of H-score  
(see Figure~\ref{fig:measures} for a visualization of the comparison). 
Hence, R-score indeed represents
a valid alternative metric with the additional advantage of computational efficiency, since
it does not require the computation of the induced facts $I(\mathcal{R}',\mathcal{F})$.

Also Herbrand accuracy and accuracy provide similar results. Observe that these two measures are not meaningful in our settings since they yield always very high performances.
Note that precision and H-score are very close for AMIE+, Neural-LP, and NTP, but not for FOIL.
This could be explained by the fact that the training of the former systems 
maximizes functions that are similar to precision, 
while FOIL uses heuristics to produce the rules that induce the maximum number of facts in the training set and minimum number of facts not in the training set.
 The great discrepancy between the two measures with FOIL means that the rules it learns do not produce many false facts but only a subset of the facts induced by the original rules.
For AMIE+ instead, since precision and H-score are similar, we have that its rules  produce most of the consequences of the original rules and, thanks to the good performance, they do not produce too many false facts.
Considering Neural-LP and NTP the two measures are also very similar, but very low: their rules produce most of the positive examples but also a lot of false facts.
In the following subsections we report H-score, but the R-score results we computed for comparison showed the same trends.

\begin{table}[h!]
    \centering
    \begin{tabular}{lcccc}
    \toprule
    & FOIL & AMIE+ & Neural-LP & NTP \\
     \midrule
    H-accuracy & 0.9873 &	0.8498 & 	0.9850 &	0.9221 \\
    Accuracy & 0.9872 &	0.8494 &	0.9849 &	0.9219\\
    F1-score &  0.2151 &	0.3031 &	0.1621 & 	0.1125 \\
    H-score & 0.1520 &	0.2321 &	0.1025 &	0.0728 \\
    Precision & 0.5963 &	0.2982 &	0.1687 &	0.1021 \\ 
    Recall & 0.2264 &	0.7311 &	0.2433 &	0.3921 \\
    R-score & 0.2728 &	0.3350  & 0.1906 & 0.1811 \\
    \bottomrule
    \end{tabular}
    \caption{Impact of different metrics, each one averaged on 120 datasets with uniformly distributed categories $\in$ \{CHAIN, RDG, DRDG\}, sizes $\in$ \{XS,S\}, and graph depths $\in$ \{2,3\}; $\nowa=0.3$, $\nnoisemiss=0.2$, $\nnoiseadd=0.1$.}
    \label{tab:result_1}
\end{table}

\subsection{Impact of Missing Consequences and Noise}\label{sec:exp_noise}
In this experiment, we evaluated the performance of the systems in the presence of complete information, incomplete information, and incomplete information with noise. This was performed analyzing the impact of the different parameters given in \tool: $\nowa$, $\nnoiseadd$, and $\nnoisemiss$. 
The results are reported in Table \ref{tab:results_2}. 
The noise parameters are defines as follows\footnote{the set memberships are intended to mean ``uniformly distributed over''.}: 
{\it complete} datasets $\nowa=0$, $\nnoisemiss=0$, and $\nnoiseadd=0$, 
{\it incomplete} datasets $\nowa\in\{0.2,0.3,0.4\}$, $\nnoisemiss=0$, and $\nnoiseadd=0$, and 
{\it incomplete + noise} datasets $\nowa\in$ \\$\{0.2,0.3, 0.4\}$, $\nnoisemiss\in\{0.15,0.3\}$, and $\nnoiseadd\in\{0.2,0.3\}$. 
Moreover, in order to give an impression of some of the datasets considered in existing evaluations, we included one manually created dataset, EVEN, inspired by the corresponding dataset used in  \cite{EGre-jair18:learning-explanatory-rules}\footnote{In our version, 
$\expred{even}(X) \ass \expred{even}(Z),\expred{succ}(Z,Y),\expred{succ}(Y,X)$ 
is the only rule,
and the input facts are such that we also have an  accuracy of $1$ if the symmetric rule $\expred{even}(Z) \ass \expred{even}(X), \expred{succ}(Z,Y),\expred{succ}(Y,X)$ is learned (using the original fact set it would be $0$). AMIE+ and Neural-LP do not support unary predicates, present in EVEN.}, which contains complete information.
We notice that FOIL shows a good performance if the information is exact and complete while showing decreasing performance in more noisy scenarios. 
This is a result of the assumptions FOIL is based upon: it 
assumes negative examples to be given in addition in order to guide rule learning and, in particular, missing facts to be false 
(see Section 4.1 in \cite{Quinlan-ML90:foil}).
AMIE+ seems to perform constant on average, showing
robustness to noise and incomplete data in all the datasets.
Neural-LP and NTP seem to be robust to noise and incomplete data, not showing changes in performance while adding more noise and uncertainty.
\begin{table}[h!]
    \centering
    \small
    \begin{tabular}{lcccc}
    \toprule
    &\multirow{2}{*}{EVEN} &  \multirow{2}{*}{Complete} & \multirow{2}{*}{Incomplete}& Incomplete \\& & & & + Noise \\
     \midrule
     FOIL & 1.0 & 0.4053 &	0.1919 &	0.0849 \\
     AMIE+ & - &  0.2021 &	0.2098 &	0.2075 \\
     Neural-LP&  - & 0.0633 &	0.0692 &	0.0649 \\
     NTP  & 1.0 & 0.0482 &	0.0617 &	0.0574 \\
     \bottomrule
    \end{tabular}
    \caption{Effect of missing consequences and noise on 144 datasets. Each H-score value is averaged on 48 datasets, with uniformly distributed categories $\in$ \{RDG, DRDG\}, sizes $\in$ \{XS,S\}, and graph depths $\in$ \{2,3\}.}
    \label{tab:results_2}
\end{table}
\subsection{Impact of Dependencies Between Rules}\label{sec:rule-dep-exps}
In this experiment, we analyze the impact of the kind of the dependencies between rules.
The results are reported in Table \ref{tab:result_3}. As expected, the systems perform very different depending on the datasets' rule categories. 
We notice that the systems perform better on the Chain datasets while only learning partially RDG and DRDG rules, meaning that the available rule learning systems are not yet able to capture complex rule set structures.

Our results also confirm the system descriptions w.r.t.\ the rules they support. 
For instance, AMIE+ does not consider reflexive rules and requires rules to be connected (i.e., every atom must share an argument with each of the other atoms of the rule) and closed (all variables appear at least twice).
%
And, by chance, the Chain datasets in \tool-v0 more often satisfy these conditions than the other datasets (this is not true in general: our generator produces comprehensive datasets that do not necessarily satisfy this property).

Nevertheless, rules that are not fully supported can still be recognized partially. We have seen this by analyzing the rules learned by Neural-LP, which only supports chain rules of form \eqref{eq:chainrule}.

NTP also performs better on Chain datasets, but the discrepancy with the other types of datasets is not substantial. This can be explained by the fact that we provided all necessary templates 
(for more details about the system requirements see \cite{RoR-NIPS17}). 

We cannot draw significant conclusions for FOIL given its unstable behaviour regarding the dataset type.

In conclusion, our experiments demonstrate 
the importance of considering 
datasets with different kinds of rules sets and of considering different measures of performance to be able to fully understand the weaknesses and strengths of a rule learning system.

\begin{table}[h!]
    \centering
    \begin{tabular}{lccc}
    \toprule
      & CHAIN &  RDG & DRDG  \\
     \midrule
     FOIL & 0.2024 &	0.0877 &	0.1633     \\
     AMIE+ &  0.3395 &	0.2275 &	0.1293 \\
     Neural-LP & 0.1291 &	0.1050 &	0.0734 \\
     NTP  & 0.1239 &	0.0538 &	0.0368  \\
     \bottomrule
    \end{tabular}
     \caption{Impact of dataset category. H-score averaged on 40 datasets. 
    Datasets as in Section~\ref{sec:exp1}.
    }
    \label{tab:result_3}
\end{table}
\subsection{Scalability: Impact of Dataset Size}
\begin{table}[h!]
    \centering
    \begin{tabular}{lcccccc}
    \toprule
      && XS-2 &  XS-3 && S-2 & S-3  \\
     \midrule 
     FOIL  && 0.2815 &	0.2074 &&	0.0356 &	0.0934 \\
     AMIE+  && 0.1449 &	0.1319 &&	0.4392 &	0.2124 \\
     Neural-LP   && 0.1155 &	0.0673 &&	0.1281 &	0.0992 \\
     NTP && 0.1512 &	0.0432 &&	0.0652 &	0.0374 \\
     \bottomrule
    \end{tabular}
    \caption{Impact of dataset size and rule graph depth. H-score averaged on 30 datasets. Datasets as in Section~\ref{sec:exp1}.}
    \label{tab:results_4}
\end{table}
In this experiment, we analyzed the impact of the dataset size considering four different size-depth combinations:  
the results for XS-2, XS-3, S-2, and S-3 datasets are reported in Table~\ref{tab:results_4}.
We can observe that FOIL is not scalable, since there is a 20\% performance gap from the XS-dataset to the S-dataset. Although it does not seem to be influenced by the rules dependency tree depth, showing support to nested rules. 
AMIE+ seemingly shows constant performance and thus scalability. We can observe that there is a noticeable decrease of performance if we increase the depth of the rule dependency graphs.
Neural-LP and NTP are robust to noise and incomplete data but NTP is not scalable yielding good accuracy only on the very small and simple instances (XS-2), while
Neural-LP seems to be more scalable (we cannot see a decrease of performance, augmenting the size of the dataset) but does not support nested rules.

\section{Conclusions and Future Work}\label{sec:conclusions}


In this paper, we have presented \tool, a system for generating datasets for rule learning and for evaluating rule learning systems.
Our experiments on new, generated datasets have shown that it is very important to have diverse datasets that consider several rule types separately, different sizes, different amount and type of noise and to perform the evaluation using different measures of performance. 
With our datasets and evaluation tool we provide these capabilities allowing to fully understand the weaknesses and strengths of a rule learning system.

There are various directions for future work. 
The dataset generation can be extended to more expressive logics
(with negation, existential quantification in rule heads, functions, etc.) 
including probabilistic inference that would allow to evaluate methods that learn probabilistic rules. 
Another possibility is to increase the probability to generate special predicate types: transitive predicates, predicates that admit only disjoint combinations of constants (e.g., the relation between a person and their SSN), or functions.

\bibliographystyle{aaai}

\clearpage
\newpage
\appendix 
\setcounter{secnumdepth}{1}

\section{System Configurations}
All the systems have the same computational restrictions (i.e. CPU, memory, time limit, etc.). The reader can find all the details (scripts etc.) in the \tool GitHub repository.

\subsection{FOIL}
\begin{itemize}
\item {\bf Paper:} Learning logical definitions from relations. Machine Learning, 5:239–266, 1990. \url{https://www.semanticscholar.org/paper/Learning-logical-definitions-from-relations-Quinlan/554f3b32b956035fbfabba730c6f0300d6955dce}
\item {\bf Source Code:} \url{http://www.rulequest.com/Personal/} or \url{http://www.cs.cmu.edu/afs/cs/project/ai-repository/ai/areas/learning/systems/foil/foil6/0.html} ,
Version: 6
\item {\bf Running configuration:}
\begin{lstlisting}[basicstyle=\ttfamily\scriptsize,showstringspaces=false]
 \$SYSDIR/\$SYSTEM/FOIL/./foil6 -v0 -n -m 200000
    < \$PREPROCESSINGFOLDER\$FILENAME.d
 \end{lstlisting}
\texttt{-m 200000}: used when the max tuples are exceeded
\item {\bf Parameter for accepting the rules:} NA -- all the rules are accepted
\end{itemize}

\subsection{Amie+}
\begin{itemize}
\item {\bf Paper:} Fast Rule Mining in Ontological Knowledge Bases with AMIE+. Luis Galárraga, Christina Teflioudi, Fabian Suchanek, Katja Hose. VLDB Journal 2015. 
\url{https://suchanek.name/work/publications/vldbj2015.pdf}
\item {\bf Source Code:} \url{https://www.mpi-inf.mpg.de/departments/databases-and-information-systems/research/yago-naga/amie/} , Version of 2015-08-26
\item {\bf Running configuration:}
\begin{lstlisting}[basicstyle=\ttfamily\scriptsize,showstringspaces=false]
java -jar \$SYSDIR/\$SYSTEM/amie\_plus.jar 
    -mins 3 -minis 3 -minpca 0.25 
    -oute \$DATA/\$SYSTEM/\$NAME/train.txt 
    > \$DIR/../output/binary/\$SYSTEM/\$NAME/results.txt
\end{lstlisting}
\item {\bf Parameter for accepting the rules:} learned using grid-search $=0.7$ -- all the rules with PCA Confidence $> 0.7$ are accepted
\end{itemize}

\subsection{Neural-LP}
\begin{itemize}
\item {\bf Paper:} Differentiable Learning of Logical Rules for Knowledge Base Reasoning.Fan Yang, Zhilin Yang, William W. Cohen. NIPS 2017. \url{https://arxiv.org/abs/1702.08367}
\item {\bf Source Code:} \url{https://github.com/fanyangxyz/Neural-LP}
\item {\bf Running configuration:}
\begin{lstlisting}[basicstyle=\ttfamily\scriptsize,showstringspaces=false]
python \$SYSDIR/\$SYSTEM/src/main.py 
    --datadir=\$DATA/\$SYSTEM/\$NAME
    --exp\_dir=$DIR/../output/binary/$SYSTEM 
    --exp\_name=\$NAME 
    > \$DIR/../output/binary/\$SYSTEM/\$NAME/log.txt 
\end{lstlisting}
\item {\bf Parameter for accepting the rules:} learned using grid-search $=0.6$ -- all the rules with ri-normalized prob $> 0.6$ are accepted
\end{itemize}

\subsection{Neural-theorem prover (ntp)}
\begin{itemize}
\item {\bf Paper:} End-to-end Differentiable Proving. Tim Rocktaeschel and Sebastian Riedel. NIPS 2017. \url{http://papers.nips.cc/paper/6969-end-to-end-differentiable-proving}
\item {\bf Source Code:} \url{https://github.com/uclmr/ntp}
\item {\bf Running configuration:}
\begin{lstlisting}[basicstyle=\ttfamily\scriptsize,showstringspaces=false]
python \$SYSDIR/\$SYSTEM/ntp/experiments/learn.py
    \$DATA/\$SYSTEM/\$NAME/run.conf 
    > \$DIR/../output/binary/\$SYSTEM/\$NAME/log.txt
\end{lstlisting}
\item {\bf Parameter for accepting the rules:} learned using grid-search $=0.0$ -- all the rules are accepted
\end{itemize}
\begin{lstlisting}[basicstyle=\ttfamily\scriptsize,showstringspaces=false]
{
    "data": {
        "kb": "$DATAPATH/$TRAIN.nl",
        "templates": "$DATAPATH/rules.nlt"
    },
    "meta": {
        "parent": "$SYSTEMSPATH/conf/default.conf",
        "test_graph_creation": False,
        "experiment_prefix": "$NAME",
        "test_set": "$TEST",
        "result_file": "$OUTPUTPATH/results.tsv",
        "debug": False
    },
    "training": {
        "num_epochs": 100,
        "report_interval": 10,
        "pos_per_batch": 10,
        "neg_per_pos": 1,
        "optimizer": "Adam",
        "learning_rate": 0.001,
        "sampling_scheme": "all",
        "init": None, # xavier initialization
        "clip": (-1.0, 1.0)
    },
    "model": {
        "input_size": 100,
        "k_max": 10,
        "name": "???",
        "neural_link_predictor": "ComplEx",
        "l2": 0.01, # 0.01 # 0.0001
        "keep_prob": 0.7
    }
}
\end{lstlisting}

\end{document}